\documentclass[runningheads]{llncs}

\usepackage{eccv}

\usepackage{eccvabbrv}

\usepackage{graphicx}
\usepackage{booktabs}
\usepackage{multirow}

\usepackage{tcolorbox}
\usepackage{xcolor}
\usepackage{caption}
\usepackage[accsupp]{axessibility}  
\usepackage[export]{adjustbox}
\usepackage{subcaption}

\usepackage{hyperref}

\usepackage{orcidlink}

\begin{document}

\title{Improving Spatial-Temporal Reasoning in Video-Language Models with Structured Video Prompting} 

\titlerunning{Structured Video Prompting}

\author{Sadegh Mohammadian\inst{1}\orcidlink{0009-0005-4628-1594}}

\authorrunning{S. Mohammadian}

\institute{Sharif University of Technology\\
\email{sadegh803211@gmail.com}}

\maketitle

\begin{abstract}
Video-language models (VLMs) remain brittle on tasks that require tracking events over time and grounding answers in specific spatial regions. We propose that part of this limitation can be addressed through better organization of visual evidence at inference time. We introduce \emph{structured video prompting}, a training-free inference-time method that augments the input video with lightweight \emph{spatial structure} and \emph{temporal structure}, providing explicit anchors for organizing evidence across space and time without changing model weights or decoding and without altering the
question prompt in the main comparison. We evaluate this approach on two complementary video benchmarks and two open video-language models. Across these settings, structured inputs improve performance in several cases, with gains varying by model and task. Our findings suggest that some failures of VLMs arise not only from reasoning capacity, but also from how video evidence is presented at inference time. These results highlight structured video prompting as a simple and practical direction for improving video understanding.
\end{abstract}

\begin{figure*}[!t]
    \centering
    \includegraphics[width=\textwidth]{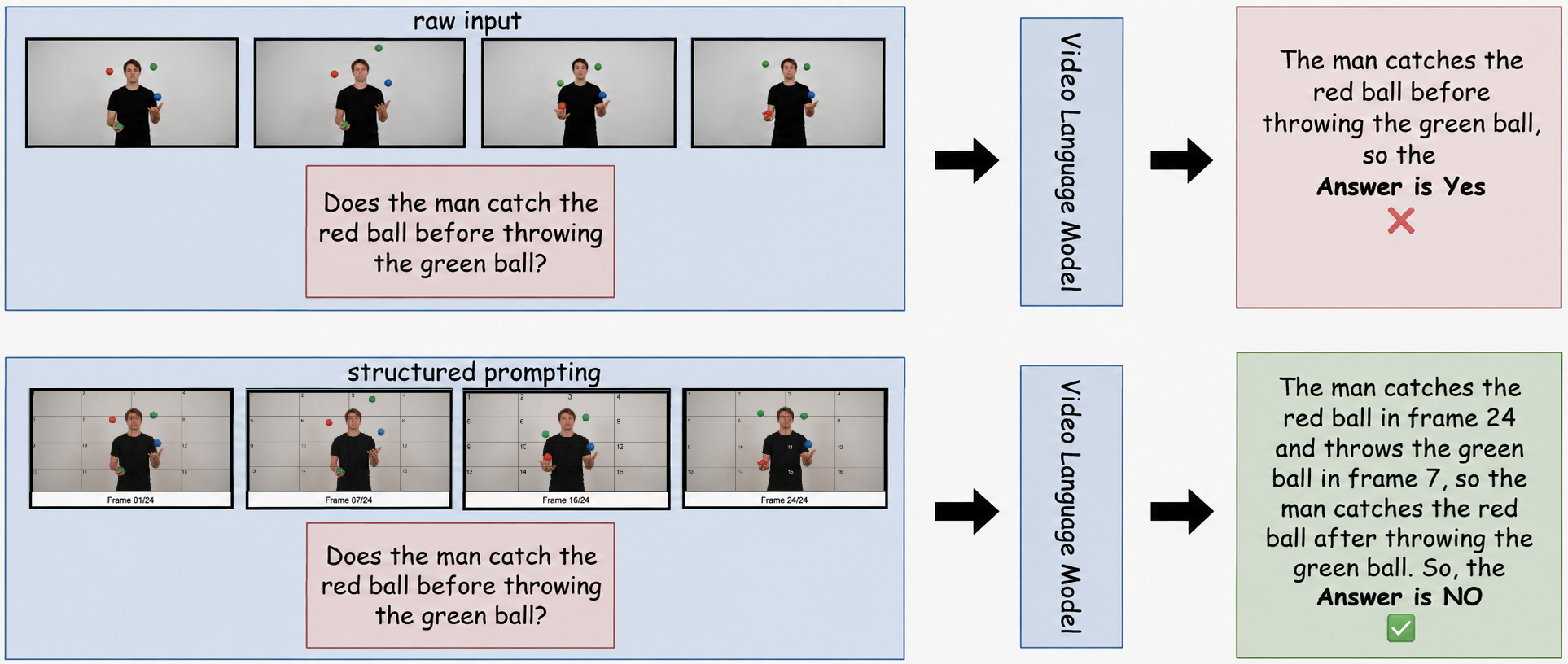}
    \vspace{-1mm}
    \caption{Overview of our method. Raw video input (top) can lead to incorrect reasoning when spatial and temporal evidence is weakly organized. Structured video prompting (bottom) adds explicit spatial and temporal anchors to the same video, helping the model identify relevant evidence and answer correctly.}
    \vspace{-2mm}
    \label{fig:method}
\end{figure*}
\section{Introduction}

Video-language models (VLMs) have recently shown strong performance on a wide range of multimodal understanding tasks~\cite{bai2025qwen25vl,zhu2025internvl3}. However, reliable video reasoning remains challenging. Even when the relevant evidence is visually present, current models often struggle with questions that require identifying event order, grounding answers in precise spatial regions, or connecting evidence across multiple moments in a video~\cite{fu2024videomme,cai2024temporalbench}. These failures suggest that errors in video understanding are not only a matter of missing knowledge or limited reasoning capacity, but also of how visual evidence is organized and presented to the model.

A possible reason is that video inputs are typically provided to VLMs as weakly structured frame sequences. While the model receives multiple frames, the spatial and temporal organization of the evidence is often implicit: temporal position is not explicitly marked, spatial references inside each frame are unlabeled, and the model must infer from raw visual content alone where relevant objects are, when they appear, and what happens to them over time. As a result, models can struggle to find the right visual evidence, relate objects to the correct moments, and track how events unfold across frames. This can make reasoning brittle, especially for tasks that require cross-frame comparison, localized grounding, or careful evidence tracking over time.

In this paper, we study a simple alternative: instead of changing the model, we modify the \emph{video input}. We build on a line of work on \emph{image-based} vision-language reasoning that analyzes model failures through the lens of the \emph{binding problem}~\cite{campbell2024bindingproblem}, arguing that models often struggle to correctly associate attributes, objects, and relations in complex visual scenes. As a lightweight solution in the image setting, VISER proposes adding simple visual structure to the input to support more organized reasoning~\cite{izadi2025viser}. Follow-up analysis further suggests that external cues such as symbols and grid lines can induce latent grounding identifiers that improve cross-modal alignment and help the model focus on relevant information~\cite{hasani2025groundingids}. Motivated by these findings, we investigate whether similar ideas can be extended from image-language models to \emph{video}-language models by adding explicit spatial-temporal scaffolding to video frames. Our method can be viewed as a lightweight inference-time aid that encourages more deliberate, stepwise reasoning over space and time by providing explicit spatial and temporal anchors for organizing evidence. It is training-free and implemented as a preprocessing step. Given an input video, we overlay a light spatial grid with cell identifiers and add temporal indicators such as frame indices in a bottom band. These modifications aim to make spatial regions and temporal positions more explicit while preserving the original visual content.

We evaluate this idea on subsets of the validation splits of \textbf{TemporalBench} and \textbf{MMSI-Video-Bench} using two open models, \textbf{Qwen2.5-VL-7B} and \textbf{InternVL3.5-8B}\cite{cai2024temporalbench,bai2025qwen25vl,
wang2025internvl35,lin2025mmsi}. These datasets let us study the effect of structured video prompting under complementary settings: temporal understanding and spatial reasoning. For each benchmark, we compare a standard baseline input against a structured version of the same video, allowing us to isolate the effect of visual input organization without fine-tuning, additional supervision, or changes to decoding.

Our results show that explicit input structure is a promising inference-time aid for video understanding. Across the evaluated subsets, structured videos improve performance in several cases, with gains depending on the model and benchmark. These findings suggest that some reasoning failures in VLMs stem not only from model limitations, but also from the way visual evidence is presented at inference time. More broadly, our study highlights structured video prompting as a simple and practical direction for improving video-language models.

\section{Related Work}

\paragraph{Video-language models and video reasoning benchmarks.}
Recent video-language models have made rapid progress in multimodal understanding, with systems such as Video-LLaMA and Video-ChatGPT establishing early open baselines for video-grounded language interaction, and newer general-purpose models such as Qwen2.5-VL and InternVL3.5 providing increasingly strong support for video understanding~\cite{zhang2023videollama,maaz2024videochatgpt,bai2025qwen25vl,wang2025internvl35}. At the same time, several recent benchmarks show that reliable video reasoning remains unsolved. Video-MME highlights broad weaknesses in video analysis for multimodal models~\cite{fu2024videomme}, TemporalBench focuses on fine-grained temporal understanding such as event order, motion, and temporal change~\cite{cai2024temporalbench}, and MMSI-Video-Bench evaluates video-based spatial intelligence~\cite{lin2025mmsi}. Together, these benchmarks suggest that current systems still struggle with complementary aspects of video understanding, including temporal confusion and weak spatial grounding.

\paragraph{Binding, grounding, and structured visual inputs.}
A related line of work studies failures of vision-language models through the lens of the \emph{binding problem}, arguing that models often struggle to correctly associate objects, attributes, and relations in complex scenes~\cite{campbell2024bindingproblem}. In the image setting, VISER shows that adding lightweight visual structure to the input can improve reasoning by making the organization of visual evidence more explicit~\cite{izadi2025viser}. Follow-up analysis further suggests that external cues such as symbols and grid lines can induce latent grounding identifiers that improve cross-modal alignment and help the model focus on relevant information~\cite{hasani2025groundingids}. Related mechanistic studies further show that model computation can be sensitive to input organization: visual layout and spatial composition affect numerical representations in vision-language models, while test-time decomposition into simpler subproblems can improve otherwise capacity-limited reasoning in language models~\cite{Hasani_2026_CVPR,hasani-etal-2026-mechanistic}. These works motivate the broader hypothesis that some multimodal reasoning failures arise not only from model limitations, but also from how visual evidence is externally structured at inference time.

\paragraph{Model-centric improvements to video reasoning.}
Most efforts to improve temporal or spatial reasoning in video-language models are \emph{model-centric}, intervening in architecture, training, attention, or decoding. Some methods explicitly modify visual-temporal representations inside the model, for example by adding timestamp-aware encoders, time tokens, or temporal localization modules, as in TimeChat, VTimeLLM, LITA, and Momentor~\cite{ren2024timechat,huang2024vtimellm,huang2024lita,qian2024momentor}. Others rely on additional training data or new instruction-tuning recipes, including Tarsier, synthetic video instruction tuning, and LLaVA-NeXT-Interleave~\cite{wang2024tarsier,zhang2024videoinstruction,li2024llavanextinterleave}. A further line changes the reasoning or decoding pipeline itself through modular or staged inference, as in MoReVQA and Video-of-Thought~\cite{min2024morevqa,fei2024videoofthought}. While these approaches improve video reasoning by changing the model or its inference procedure, they typically require extra training, auxiliary modules, or more complex pipelines.

\paragraph{Our position.}
Our work is closest to the structured-input line, but extends it from image-language to \emph{video}-language reasoning. Rather than changing the model or using multi-stage inference, we study a simple training-free input transformation that adds lightweight spatial and temporal anchors directly to video frames. This lets us test whether some video reasoning failures stem from input presentation rather than model capacity alone.

\section{Method}

We propose \emph{structured video prompting}, a training-free inference-time method that adds lightweight spatial and temporal cues directly to the input video. Instead of modifying model parameters or decoding, we transform each video before passing it to the video-language model. The goal is to make the spatial and temporal organization of visual evidence more explicit, which may help the model localize relevant content and reason across time more reliably.

\subsection{Structured Video Transformation}

Given an input video $V=\{f_1, f_2, \dots, f_T\}$ with $T$ frames, we construct a structured version
\begin{equation}
\tilde{V} = \mathcal{T}(V),
\end{equation}
where $\mathcal{T}$ is a deterministic preprocessing function applied frame-wise, conditioned on the frame index $t$ and total length $T$. For each frame $f_t$, we add two types of visual scaffolding.

\paragraph{Spatial structure.}
We overlay a light $4 \times 4$ spatial grid on the frame and assign each cell a unique identifier. This divides the frame into 16 cells indexed in row-major order. Thin grid lines separate adjacent cells, and each cell number is rendered near its top-left corner. This provides explicit spatial references that may help the model associate visual content with localized regions and better ground question-relevant evidence.

\paragraph{Temporal structure.}
We append a small auxiliary band below each frame that displays the temporal index of the current frame, e.g., \texttt{Frame $t/T$}. This makes temporal position directly visible in the video itself rather than leaving it implicit in the frame sequence, which may help the model distinguish evidence from different moments and track how events unfold over time.

The resulting structured frame preserves the original visual content while augmenting it with explicit spatial and temporal references. To keep the intervention lightweight, grid lines are thin, labels are small, and the temporal band is placed outside the original frame region. Figure~\ref{fig:method} illustrates the transformation from the original video to the structured video.

Our method tests whether explicit spatial and temporal anchors can improve video reasoning by helping models localize, distinguish, and connect visual evidence across frames.

\subsection{Inference Setup}

Let $q$ denote a question about video $V$. In the baseline setting, the model receives the original video $V$ and the question $q$. In the structured setting, the model receives the transformed video $\tilde{V}$ with the same question. We do not modify the prompt between the two settings. Formally, for a video-language model $\mathcal{M}$, we compare
\begin{equation}
y_{\text{base}} = \mathcal{M}(V, q), \qquad
y_{\text{struct}} = \mathcal{M}(\tilde{V}, q),
\end{equation}
where $y_{\text{base}}$ and $y_{\text{struct}}$ are the model predictions under the baseline and structured conditions, respectively. Unless otherwise stated, all inference settings are kept fixed across the two conditions, including decoding parameters and video sampling settings, so that the comparison isolates the effect of visual input structure.

\section{Experiments}
\label{sec:experiments}

\begin{table*}[!t]
\centering
\small

\begin{subtable}[t]{0.485\textwidth}
\centering
\caption{Main results on the evaluated subsets. Accuracy (\%).}
\label{tab:main_results}
\resizebox{\linewidth}{!}{%
\begin{tabular}{@{}lcccc@{}}
\toprule
\multirow{2}{*}{Benchmark}
& \multicolumn{2}{c}{Qwen2.5-VL-7B}
& \multicolumn{2}{c}{InternVL3.5-8B} \\
\cmidrule(lr){2-3}
\cmidrule(lr){4-5}
& Base & Struct. & Base & Struct. \\
\midrule
TemporalBench
& 60.75 & \textbf{63.00}
& 63.25 & \textbf{66.00} \\

MMSI-Video-Bench
& 28.25 & \textbf{33.25}
& 31.75 & \textbf{34.00} \\
\bottomrule
\end{tabular}%
}
\end{subtable}
\hfill
\begin{subtable}[t]{0.485\textwidth}
\centering
\caption{Category-level results on MMSI-Video-Bench. Accuracy (\%).}
\label{tab:mmsi_categories}
\resizebox{\linewidth}{!}{%
\begin{tabular}{@{}lcccc@{}}
\toprule
\multirow{2}{*}{Category}
& \multicolumn{2}{c}{Qwen2.5-VL-7B}
& \multicolumn{2}{c}{InternVL3.5-8B} \\
\cmidrule(lr){2-3}
\cmidrule(lr){4-5}
& Base & Struct. & Base & Struct. \\
\midrule
Spatial Construction
& 27.74 & \textbf{34.19}
& 32.26 & \textbf{33.55} \\

Motion Understanding
& 23.47 & \textbf{27.55}
& \textbf{38.78} & 36.73 \\

Cross-Video Reasoning
& 28.17 & \textbf{36.62}
& 32.39 & \textbf{40.85} \\

Planning
& 37.74 & 37.74
& 22.64 & \textbf{24.53} \\

Prediction
& 30.43 & 30.43
& 17.39 & \textbf{26.09} \\
\bottomrule
\end{tabular}%
}
\end{subtable}

\vspace{-2mm}
\end{table*}

\subsection{Benchmarks}
\label{sec:benchmarks}

We evaluate structured video prompting on two complementary video
reasoning benchmarks. \textbf{TemporalBench}~\cite{cai2024temporalbench}
evaluates fine-grained temporal understanding, including event order,
motion, and changes over time. \textbf{MMSI-Video-Bench}~\cite{lin2025mmsi}
evaluates video-based spatial intelligence, including spatial
construction, motion understanding, cross-video reasoning, planning,
and prediction.

To keep the evaluation computationally manageable while maintaining a
controlled comparison, we use fixed subsets of the validation splits.
Each subset is sampled once and reused across all models and input
conditions.

For \textbf{TemporalBench}, we evaluate on \textbf{400 validation
questions} drawn from \textbf{392 unique videos}. For
\textbf{MMSI-Video-Bench}, we evaluate on \textbf{400 validation
questions} associated with \textbf{471 unique videos} and
\textbf{103 reference images}.

The MMSI-Video-Bench subset covers five categories:
\textit{Spatial Construction} (155 questions),
\textit{Motion Understanding} (98),
\textit{Cross-Video Reasoning} (71),
\textit{Planning} (53), and
\textit{Prediction} (23).

\subsection{Models and Inference Setup}
\label{sec:models_setup}

We evaluate two open video-language models:
\textbf{Qwen2.5-VL-7B}~\cite{bai2025qwen25vl} and
\textbf{InternVL3.5-8B}~\cite{wang2025internvl35}.
For the main experiments, we compare two inference-time conditions:

\begin{itemize}
    \item \textbf{Baseline}: the original video is provided to the model.
    \item \textbf{Structured}: the same video is transformed using
    structured video prompting.
\end{itemize}

In the structured condition, each sampled frame is augmented with a
light $4\times4$ spatial grid, cell identifiers, and a temporal band
showing the frame index. The question prompt is kept identical between
the baseline and structured conditions.

\paragraph{Implementation details.}
For Qwen2.5-VL-7B, videos are sampled using \texttt{fps=1}. For
InternVL3.5-8B, we set \texttt{max\_num\_frames=64}. All decoding
parameters and video-sampling settings are held fixed between the
baseline and structured conditions. Each benchmark subset is sampled
once using a fixed random seed and reused across models and conditions.
The only experiment that changes the language prompt is the text-only
temporal ablation described in Sec.~\ref{sec:ablation}.

\subsection{Evaluation Protocol}
\label{sec:evaluation_protocol}

We report exact-match accuracy for each benchmark and model. The main
comparison changes only the visual rendering of the video, isolating
the effect of spatial-temporal organization. The ablation also includes
a text-only temporal instruction to compare visual frame identifiers
with language-based temporal cues.

\subsection{Main Results}
\label{sec:main_results}

Table~\ref{tab:main_results} presents the main results. Across both
benchmarks and both models, structured video prompting improves
performance over the baseline.

On \textbf{TemporalBench}, Qwen2.5-VL-7B improves from 60.75\% to
63.00\%, corresponding to a gain of 2.25 percentage points.
InternVL3.5-8B improves from 63.25\% to 66.00\%, a gain of 2.75 points.

On \textbf{MMSI-Video-Bench}, Qwen2.5-VL-7B improves from 28.25\% to
33.25\%, yielding the largest overall gain of 5.00 points.
InternVL3.5-8B improves from 31.75\% to 34.00\%, a gain of 2.25 points.
These results indicate that lightweight visual structure can serve as
an effective inference-time aid for both temporal and spatial video
reasoning.

The magnitude of the improvement varies across benchmarks and models.
For Qwen2.5-VL-7B, the larger improvement on MMSI-Video-Bench suggests
that explicit spatial anchors are particularly beneficial. For
InternVL3.5-8B, improvements appear on both benchmarks, with a slightly
larger gain on TemporalBench, indicating that visible temporal anchors
can also help organize evidence across frames.

\subsection{Category-Level Results on MMSI-Video-Bench}
\label{sec:category_results}

Table~\ref{tab:mmsi_categories} shows that structured prompting yields
the largest gains in \textit{Cross-Video Reasoning}, improving
Qwen2.5-VL-7B from 28.17
32.39
\textit{Spatial Construction}, while InternVL3.5-8B gains most in
\textit{Prediction}. Other categories show smaller or inconsistent
changes. Overall, structured prompting is most effective for spatial
grounding and cross-frame evidence organization.

\subsection{Component Ablation}
\label{sec:ablation}

To isolate the contribution of each part of structured video prompting,
we conduct an ablation study using Qwen2.5-VL-7B on the same
400-question MMSI-Video-Bench subset. We evaluate five input
conditions:

\begin{itemize}
    \item \textbf{Base}: the original video without additional structure.
    \item \textbf{Grid only}: the numbered spatial grid is added without
    visual frame identifiers.
    \item \textbf{Visual frame IDs only}: visual frame-index indicators
    are added without the spatial grid.
    \item \textbf{Text-only frame-index instruction}: the original video
    is preserved, while the language prompt explicitly describes the
    chronological ordering of sampled frames.
    \item \textbf{Structured}: the complete method combining the spatial
    grid and visual frame identifiers.
\end{itemize}

For the text-only condition, we append the following instruction to the
question prompt:

\begin{quote}
\small
\emph{The video is sampled into frames in chronological order. Treat
the first sampled frame as Frame 1 and the last sampled frame as Frame
N, where N is the total number of sampled frames. Use these frame
indices to reason about event order and temporal changes.}
\end{quote}

\begin{figure*}[!t]
    \centering

    \noindent
    \adjustbox{valign=c}{%
    \begin{minipage}{0.43\textwidth}
        \centering
        \scriptsize

        \resizebox{\linewidth}{!}{%
        \begin{tabular}{@{}lccccc@{}}
            \toprule
            Category
            & Base
            & Grid
            & Visual ID
            & Text ID
            & Struct. \\
            \midrule

            Spatial Construction
            & 27.74
            & \textbf{34.80}
            & 29.03
            & 29.68
            & 34.19 \\

            Motion Understanding
            & 23.47
            & 24.49
            & \textbf{29.49}
            & 25.51
            & 27.55 \\

            Cross-Video Reasoning
            & 28.17
            & 28.17
            & 29.58
            & 30.99
            & \textbf{36.62} \\

            Planning
            & 37.74
            & 37.74
            & 37.74
            & 37.74
            & 37.74 \\

            Prediction
            & 30.43
            & 30.43
            & 30.43
            & 30.43
            & 30.43 \\

            \midrule
            \textbf{Overall}
            & 28.25
            & 31.24
            & 30.47
            & 30.00
            & \textbf{33.25} \\

            \bottomrule
        \end{tabular}%
        }
    \end{minipage}
    }
    \hfill
    \adjustbox{valign=c}{%
    \begin{minipage}{0.54\textwidth}
        \centering
        \includegraphics[
            width=\linewidth
        ]{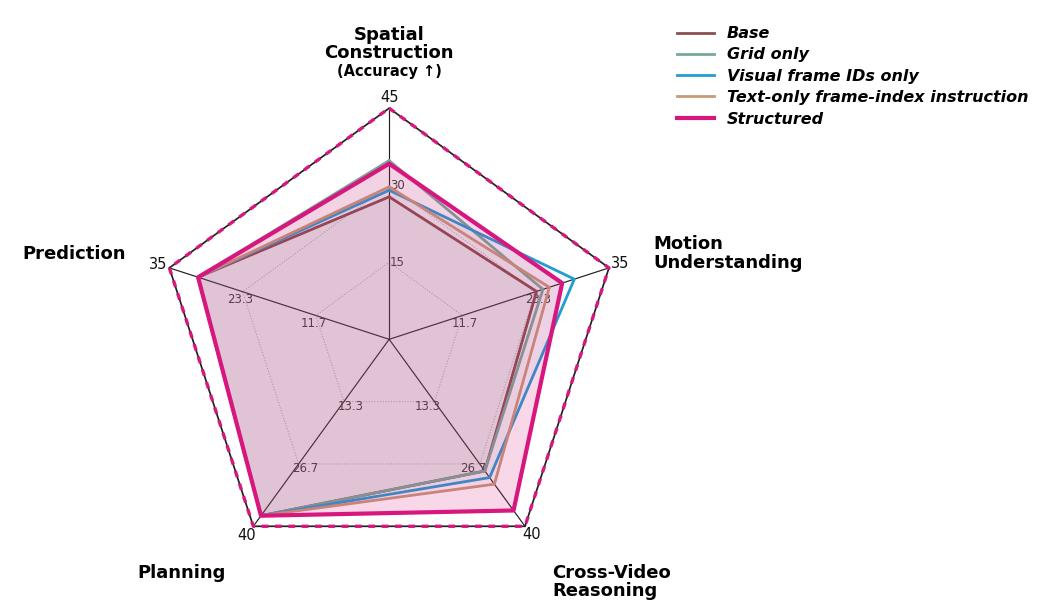}
    \end{minipage}
    }

    \vspace{1mm}

    \noindent
    \begin{minipage}[t]{0.43\textwidth}
        \centering
        \textbf{(a) Exact accuracy values (\%)}
    \end{minipage}
    \hfill
    \begin{minipage}[t]{0.54\textwidth}
        \centering
        \textbf{(b) Category-level comparison}
    \end{minipage}

    \vspace{1mm}

\caption{
Component ablation on the 400-question MMSI-Video-Bench subset using
Qwen2.5-VL-7B. \textbf{(a)} Category-level and overall accuracy.
\textbf{(b)} Radar chart across five input conditions, with each axis
independently normalized.
}

    \label{fig:ablation}
\end{figure*}

As shown in Fig.~\ref{fig:ablation}, the individual components provide
different benefits. The spatial grid primarily improves
\textit{Spatial Construction}, increasing accuracy from 27.74\% to
34.80\%. In contrast, visual frame identifiers provide the strongest
isolated result on \textit{Motion Understanding}, improving accuracy
from 23.47\% to 29.49\%.

The complete structured method achieves the highest overall accuracy,
improving from 28.25\% to 33.25\%. Its largest advantage appears in
\textit{Cross-Video Reasoning}, where accuracy increases from 28.17\%
to 36.62\%. This improvement is substantially larger than the gain from
either the grid or frame identifiers alone, suggesting that spatial and
temporal anchors provide complementary information when evidence must
be related across frames.

Performance on \textit{Planning} and \textit{Prediction} remains
unchanged across all ablation conditions. These results indicate that
structured prompting primarily assists evidence localization and
cross-frame organization rather than uniformly improving every form of
video reasoning.


\section{Conclusion}

We introduced \emph{structured video prompting}, a simple training-free method that adds explicit spatial and temporal structure directly to video inputs. Across \textbf{TemporalBench} and \textbf{MMSI-Video-Bench}, this lightweight input transformation improved performance in several settings for \textbf{Qwen2.5-VL-7B} and \textbf{InternVL3.5-8B}. These results suggest that some failures of video-language models stem not only from model limitations, but also from how visual evidence is organized at inference time. Overall, our findings highlight structured video prompting as a simple and practical direction for improving video understanding. Future work can study why this intervention works and whether it scales to more adaptive structures, longer videos, and larger models.


%
%
\bibliographystyle{splncs04}
\bibliography{main}

\clearpage
\appendix
\subsection{Qualitative Analysis}
\label{sec:qualitative}

\begin{figure*}[!t]
\centering

\begin{tcolorbox}[
    width=\textwidth,
    colback=white,
    colframe=black,
    title={Qualitative Example: Raw vs.\ Structured Video Prompting},
    fonttitle=\bfseries,
    sharp corners,
    boxrule=0.8pt
]

\centering
\includegraphics[width=0.98\textwidth]{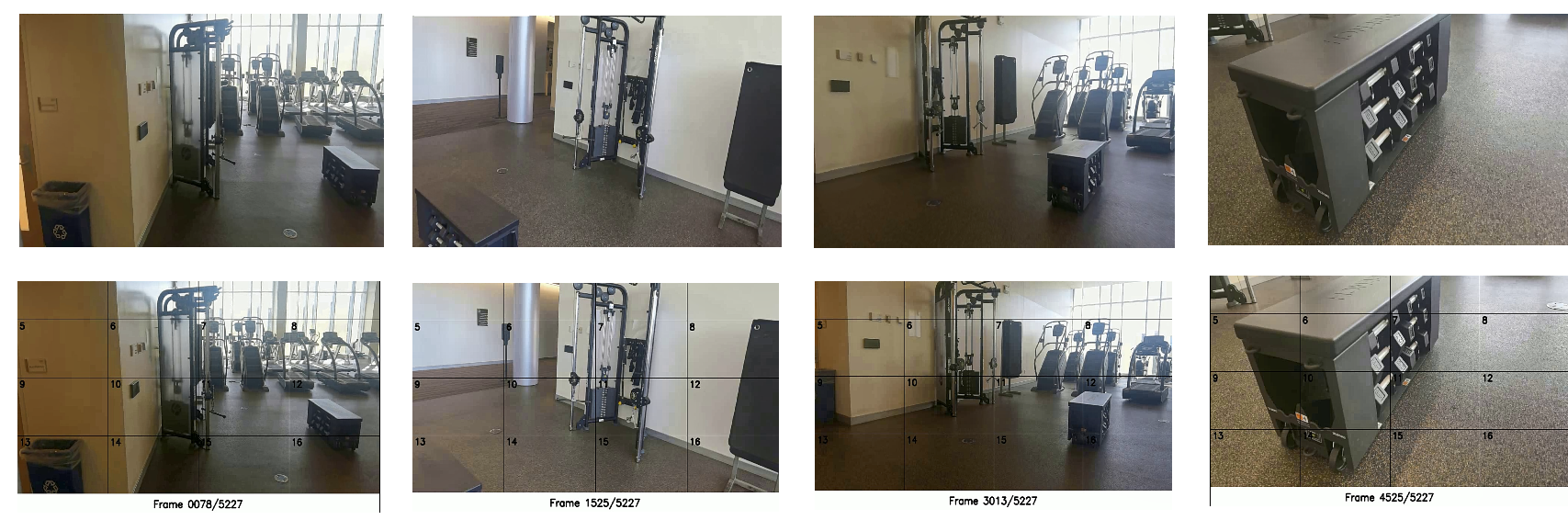}

\vspace{2mm}

\begin{tcolorbox}[
    colback=gray!8,
    colframe=black!50,
    title={Question summary},
    fonttitle=\bfseries,
    sharp corners,
    boxrule=0.5pt
]
\small
The video records a walk around a gym, and the final frame represents
the user's current position. The model must determine how the user
should leave the gym to find a friend who is wearing green and running.
The answer is selected from four candidate navigation routes.
\end{tcolorbox}

\vspace{2mm}

\noindent
\begin{minipage}[t]{0.485\textwidth}
\begin{tcolorbox}[
    colback=red!4,
    colframe=red!60!black,
    title={Raw video prediction (incorrect)},
    fonttitle=\bfseries,
    sharp corners,
    boxrule=0.5pt,
    height=39mm,
    valign=top
]
\small\ttfamily
\{"answer": "A",\\
"reason": "Rotate counterclockwise, walk to the end of the wall,
then turn left and enter."\}
\end{tcolorbox}
\end{minipage}
\hfill
\begin{minipage}[t]{0.485\textwidth}
\begin{tcolorbox}[
    colback=green!4,
    colframe=green!50!black,
    title={Structured video prediction (correct)},
    fonttitle=\bfseries,
    sharp corners,
    boxrule=0.5pt,
    height=39mm,
    valign=top
]
\small\ttfamily
\{"answer": "B",\\
"reason": "Rotate counterclockwise, walk to the end of the wall,
then turn right and enter."\}
\end{tcolorbox}
\end{minipage}

\end{tcolorbox}

\caption{
Qualitative comparison between raw and structured video prompting on
an MMSI-Video-Bench example using Qwen2.5-VL-7B. The upper and lower
rows show sampled frames from the raw and structured videos,
respectively. The raw input leads the model to select option
\textbf{A}, which is incorrect, whereas the structured input leads it
to select the correct option, \textbf{B}. Explicit spatial and temporal
anchors help the model organize evidence across the sampled frames and
infer the correct navigation direction.
}
\label{fig:qual_example}
\end{figure*}

Figure~\ref{fig:qual_example} presents a qualitative example from
MMSI-Video-Bench. The video records a walk through a gym, and the model
must infer how the user should navigate from the position shown in the
final frame.

Given the raw video, the model selects option \textbf{A}, incorrectly
concluding that the user should turn left after reaching the end of the
wall. With structured video prompting, the same model selects the
correct option, \textbf{B}. The structured frames make the spatial
layout and the temporal position of the final observation more
explicit, helping the model infer that the user should instead turn
right.

\end{document}